# Tell me a story about yourself: The words of shopping experience and self-satisfaction

Petruzzellis, L., Fronzetti Colladon, A., Visentin, M., & Chebat, J.-C.

# Tell me a story about yourself: The words of shopping experience and self-satisfaction

Petruzzellis, L., Fronzetti Colladon, A., Visentin, M., & Chebat, J.-C.[†]

**Abstract**

In this paper we investigate the verbal expression of shopping experience obtained by a sample of customers asked to freely verbalize how they felt when entering a store. Using novel tools of Text Mining and Social Network Analysis, we analyzed the interviews to understand the connection between the emotions aroused during the shopping experience, satisfaction and the way participants link these concepts to self-satisfaction and self-identity. The results show a prominent role of emotions in the discourse about the shopping experience before purchasing and an inward-looking connection to the self. Our results also suggest that modern retail environment should enhance the hedonic shopping experience in terms of fun, fantasy, moods, and emotions.

[†] This project was started before the demise of Jean-Charles Chebat and carried out in his memory

## Introduction

Shopping is much more than the simple purchase of products and services. It is a way to express the self, empower the individual identity, and live a social experience that, in turn, may enrich the value customer associate to visiting the store (Ekici et al., 2018; Picot-Coupey et al., 2021; Sirgy et al., 2016; Terblanche, 2018). The way customers experience the store is also function of the interaction between products, context and other people, which creates an enjoyable and meaningful shopping experience (Babin et al., 1994; Puccinelli et al., 2009; Leroi-Werelds et al., 2014).

Although individuals buy products mainly based on functional congruity, they engage in shopping to self-express and enhance their well-being based on self-congruity (Ekici et al., 2018; El Hedhli et al., 2021). Beyond the utilitarian dimension, the shopping experience is associated to recreation and entertainment (Guiry et al., 2006; Picot-Coupey et al., 2021). Consequently, shopping has not only to be investigated from the perspective of the utilitarian/hedonic dichotomy, but for its ability to provide individuals a chance to express their own personal identity (Guiry et al., 2006). Unfortunately, although self-expressive activities have been found to have enduring and positive effects on life satisfaction (Ryan et al., 2008), we still have limited knowledge on how an enjoyable and meaningful engagement in shopping, namely a self-expressive shopping experience, contribute to self-satisfaction (Leroi-Werelds et al., 2014; Picot-Coupey et al., 2021; Sirgy et al., 2000).

While shopping, the subjective experience generated by the congruency between the product/brand bought and the consumers' self-concept, affects customers' satisfaction towards brands, brand preferences and purchase intentions, and facilitates positive word-of-mouth and attitudes towards products (Sirgy et al., 2016). Previous research (Chebat et al., 2006; Kaltcheva et al., 2010; Sirgy et al., 2016; Sirgy et al., 2000) has found a causal link between the consumers' self-concept and shopping behaviors, which is mediated by the store atmosphere (Babin and Darden, 1995). In fact, some of the atmospheric cues shape consumer's shopping experience (Picot-Coupey et al., 2021). This, in turn, impacts the individual evaluation of products (Chebat and Michon, 2003; Morrin and Ratneshwar, 2003) as well as the recreational dimension of the shopping experience (Baker et al., 2002; Guiry et al., 2006). However, the in-store shopping experience is influenced by different cognitive and affective cues (Grewal and Roggeveen, 2020; Manthiou et al., 2020; Puccinelli et al., 2019; Visentin and Tuan, 2021).

A number of studies investigated the effect of shopping experience on life satisfaction and subjective well-being (Davis and Hodges, 2012; Ekici et al, 2018), and the relationship between the effects of self-congruity on customer satisfaction (e.g., Hosany and Martin, 2012; Ibrahim and Najjar, 2008). However, limited research (e.g., Atulkar and Kesari, 2017; Chebat et al., 2006; Sirgy et al., 2000) have looked into the extent to which the perception of the shopping experience can lead to self-satisfaction.

The main objective of the present paper is to shed light on the concept of self-satisfaction, under the assumption that consumers may undergo a form of satisfaction based on their in-store shopping experience and not based on their purchases or consumption. We offer an exploratory contribution by interviewing consumers with maximum three tattoos during their shopping experience. This choice provided participants that use body modifications to extend their reflexive self-concept and consequently are more confident with their self-identity narrative (Johnson, 2007; Kosut, 2000; Velliquette et al., 1998). The choice to limit the number of tattoos ensures that their self-identity is not excessively focused on them. After collecting the verbalization of participants' shopping experience, we conducted a textual analysis to identify the linguistic drivers and associations regarding their shopping experience. Results suggest that self-satisfaction is connected but separated from self-concept and allow us to speculate on its relationship with emotions and sensations aroused during the shopping experience.

The relevance of this study is threefold: first, by applying textual analysis, the present research provides a novel picture of the mental associations of individuals about their elaboration of the link between their self-satisfaction and their mental representation of the experience; second, this study offers empirical support to the importance of the in-store consumer experience in real brick-and-mortar contexts, mitigating the claim of big online platforms to be able to provide a higher shopping experience; third, we offer interesting new avenues for marketing research on the uncharted territory of verbal representation of an individual's shopping experience.

## Theoretical background

The shopping experience is influenced by several factors in a store, among which psychological, social and sensory cues have a relevant influence (Michon et al., 2005; Visentin and Tuan, 2021). All atmospheric elements are key to emotional states and customer satisfaction in the retail environment (Puccinelli et al., 2009). Burns and Neisner (2006) found that both cognitive evaluation and emotional reaction explain the level of satisfaction that customers experience in retail.

Traditionally rooted in Bagozzi's (1992) "appraisal – emotional response – coping" framework, the effect of perceived value on the early evaluation of the experience in the appraisal stage leads to satisfaction in the emotional response stage that determines behavioral intentions (e.g., engaging in shopping, re-patronage and word-of-mouth) in the final stage of coping (Atulkar and Kesari, 2017; Terblanche, 2018).

The experiences consumers live in a store evoke the self-concept components, namely dispositional characteristics and traits; perceptions of the ongoing or past experiences, social roles, and behaviors; self-related attitudes and affect (Ed Hedhli et al., 2021; Sirgy et al, 2016). Generally speaking, people associate their self-concept with one or more attributes, such as traits (e.g., vain, strong), behaviors/activities (e.g., choosing, shopping, thinking), physical characteristics (e.g., elegant), social categories (e.g., me, we, team), objects (e.g., brand, music), among others (Greenwald et al., 2002). Chebat et al. (2006) found that upscale shopping mall environments elicit upscale self-relevant attributes. Since people are likely to view themselves as more upscale than they actually are, consumers feel higher congruity with upscale shopping malls and evaluate the stores located in such malls more positively (Chebat et al., 2006).

Previous studies on the effects of self-congruity on shoppers' behaviors (Chebat et al., 2009; El Hedhli et al., 2017) have only considered actual self-congruity. However, this study extends prior research by conceptualizing self-satisfaction as a construct related to self-expressiveness that may lead consumers to experience positive affect while engaging in shopping.

*The influence of identity on self-satisfaction in shopping*

The self-image congruity theory has widely demonstrated that consumers define their self-concept through their product/brand choices, thus affecting their behavior, including purchase intentions, attitudes toward brands and customer satisfaction (Sirgy et al., 1997; 2016; Matzler et al., 2005; Lemon and Verhoef, 2016).

However, in retail environments customer experiences may vary according to rational, emotional, sensorial, and physical involvement, being an interplay between individual expectations and the interactions with the retail and the brand (Verhoef et al., 2009; Guiry et al., 2006). Consumers develop various impressions of their in-store experiences by assessing their array of intrinsic cues (e.g., texture, aroma) and extrinsic cues (e.g., price, brand name, packaging) (Visentin and Tuan, 2021). In addition, patrons often interact more effectively with physically attractive service providers, leading to higher consumer satisfaction and stronger purchase intentions (Ahearne et al., 1999; Soderlund and Julander, 2009). Both visual appeal and the entertainment dimension of the aesthetic response offer immediate pleasure for its own sake, irrespective of a retail environment's ability to facilitate the accomplishment of a specific shopping task (Wan and Wyer, 2015). Noteworthy, satisfaction based on emotions has been found a more robust predictor for future behavioral intentions than cognitive measures (Martin et al., 2008).

Against the common assumptions on satisfaction (e.g., Oliver, 1980; Cronin et al., 2000), customer satisfaction can be categorized as the feelings of happiness, fulfillment and pleasure towards a service provider and its services (El-Adly and Eid, 2015), based on the strong link between hedonic shopping value and satisfaction and, in turn, loyalty (Atulkar and Kesari, 2012; Babin et al., 2005; Chebat et al., 2009). This can happen during the shopping process through sensory and emotional elements thus generating self-satisfaction without the actual purchase or use. Consequently, we expect that individuals may experience some forms of excitement or arousal that galvanizes satisfaction even in absence of purchase during a shopping trip.

*Tattoos as a sign of identity*

As individuals use symbols to express something about themselves in order to social interact and express meanings, appearance affects perceptions of a person's competence, expertise, trustworthiness, personality, and intellectual capabilities (Wan and Wyer, 2015). Among people with the highest confidence with their self-concept, tattooed people stand out since tattoos are a form of marking identity (Johnson, 2007; Littel, 2003; Velliquette et al., 1998). In fact, as the body increasingly participates to the construction of an individual's self-identity, it enriches with new meanings and new statuses recognized and learned through socialization (Sweetman, 2000). Tattoos are a vehicle for human expression; they are signals of art, fashion, individuality, personal narrative, cultural tradition, group identity, individuality, freedom, and uniqueness (e.g., Burgess and Clark, 2010; Rodriguez Cano and Sams, 2010). Most of the literature maintains the association of the tattoo with the extended self, supporting that they tend to express identity narratives related to individualism (DeMello, 2000; Turner, 1999). However, tattoos also signal a connection to others, which structures perceptions and experiences (Diprose, 2005). Since they are rich in rhetoric, all these elements may shape the construction of a consumers' self-identity and empower the tattooed person in their ability to express it verbally (Kosut, 2000; Littel, 2003).

Furthermore, the number of tattoos a person has may signal different personality traits. In fact, people with more than 3 tattoos signal their belonging to tribes, subcultures or groups or their affection by a sort of mania, while people with at most 3 tattoos use these symbols as cues of fashion, style and self-confidence (Goulding et al., 2004).

*The language of the shopping experience*

Individuals tend to encode their external experiences through feelings (Kübler et al., 2020; Netzer et al., 2019; Ziemer and Kormaz, 2017). A person's preferred sensory representation system could be also evidenced from the way in which they use language (Humphreys and Wang, 2018). Literature suggests that *what* people say and *how* they say it reveal how they construe their world (e.g., Aleti et al., 2019; Berger et al., 2020). Therefore, the psychological mechanisms of this construction emerge from their writing styles, being the words used also a way to discover one's personality Humphreys and Wang, 2018; Netzer et al., 2019). In fact, there is a strict relationship between the words people use and their personality traits and identity, as well as their emotions (Hirsh and Peterson, 2009; Kosinski et al., 2013). The language is also fundamental to understand the way individuals interact with others, since people use words to understand and represent impressions of others (Berger et al., 2020; Xu and Zhang, 2018). Given that an audience is more receptive to a message arousing affective states, emotional language is more likely to be shared with others (Akpinar and Berger, 2017; Xu and Zhang, 2018).

Moreover, as the linguistic style is usually unconscious, it reveals more accurately than the content the attempt to manage individual impressions and relationships with other objects in their world (Ludwig et al., 2013). For example, during their first stages of online shopping, consumers are more likely to use abstract language, while the opposite is true for the last stages (Berger et al., 2020).

**Methodology**

In this study we focused on tattooed shoppers to uncover the relationship between the shopping experience and the concept of self-satisfaction. As explained, a sample of tattooed people can increase the likelihood that respondents have already reflected to the rhetoric of symbols building their personal identity. Thus, they can provide useful insights about self-satisfaction, when asked to verbalize their perception of a shopping experience.

We randomly intercepted 70 shoppers with at most three tattoos (Goulding et al., 2004; Velliquette et al., 1998) in a mall of a big city in South of Italy. Respondents were asked to freely verbalize how they feel when they enter a store and what galvanizes their attention during shopping. Ranging from 30 to 90 minutes, semi structured interviews were conducted with each participant. The main focus of the data collection process was to obtain data to deductively answer the study's main research question about the influence of shopping experience on self-satisfaction.

In order to evaluate the importance of the key concepts related to shopping experience and self-satisfaction, we used the Semantic Brand Score (SBS) indicator (Fronzetti Colladon, 2018), a measure of semantic importance applicable to any term/concept in a discourse. The SBS comprises three dimensions (Fronzetti Colladon, 2018): prevalence, diversity and connectivity. Prevalence represents the frequency with which a concept appears in a set of text documents (interviews in our case). The more frequently a concept is mentioned, the higher its prevalence. The second dimension, diversity, relies on the analysis of word co-occurrences, which is the textual associations of each different term in the text. It measures the heterogeneity of the words co-occurring with a term, assigning higher diversity to concepts/terms embedded in a rich and distinctive discourse (Fronzetti Colladon and Naldi, 2020). The higher the number of textual associations a term has, the more heterogeneous the semantic context in which it is used. Diversity is higher when textual associations with a concept are more diverse and is consistent with previous research showing the positive effect of a higher number of associations on concept strength. The third component, connectivity, expresses how often a term serves as an indirect link between all the other pairs of words, while considering the co-occurrence network. It reflects the embeddedness of a term/concept in a discourse and can be considered as the expression of its connective power. While a term could be frequently mentioned (high prevalence) and might have heterogeneous associations to other terms (high diversity), its concept could still be peripheral and not connected to the core of the discourse. By contrast, a concept of high importance will get more attention: it will be mentioned with high frequency, embedded in a rich discourse (diversity), and will act as a bridge across different conversation topics (connectivity). Connectivity is operationalized through the metric of weighted betweenness centrality (Brandes, 2001).

*Data analysis*

The analyses were carried out using the SBS BI web app (Fronzetti Colladon and Grippa, 2020). First, we pre-processed text data to remove stop-words (e.g., 'and'), punctuation and special characters. Then, we changed every word to lowercase and extracted stems by removing word affixes and using the NLTK Snowball Stemmer algorithm (Perkins, 2014). After pre-processing data, text documents were transformed into a semantic network where nodes are words that appear in the text. An arc exists between a pair of nodes if their corresponding words co-occurred at least once; arc weights are determined by the frequency of co-occurrence. We filtered out negligible co-occurrences, retaining only the arcs that had a minimum weight of 2. We adopted a five-word window for the determination of co-occurrences maximum range (Fronzetti Colladon, 2018).

The SBS was calculated as the sum of the standardized values of its components (Fronzetti Colladon, 2018). According to this standardization procedure, SBS scores can either be positive or negative based on the importance of a certain term.

Two academic experts in consumer behavior analyzed an initial set of keywords (Packard et al., 2018; Pennebaker et al., 2015; Rocklage et al., 2018). They met twice to select and cluster the words related to the shopping experience and the concept of (self-)satisfaction. Eight clusters emerged from the analysis, related to the concepts of: price ('prezzo' in Italian), the individual dimension including words like me/I and my/mine ('me/io' and 'mio' in Italian), pleasure ('piacere' in Italian), choice ('scelta' in Italian), sensation ('sensazione' in Italian), shopping and buying ('acquisto' in Italian), satisfaction ('soddisfazione' in Italian) and style ('stile' in Italian).

## Results

On average, the answers were of 764 words, with a type/token ration of 43%. The language used was of an average/low complexity, with 31% of words being of six letters or longer. Moreover, in relation to words expressing emotions the interview answers were positive overall, with positive emotions being about 70% more frequent than negative emotions. The answers also contained a few informal

terms – such as laughs, bad words or slang terms – indicating that the language used by the interviewees was mostly direct and informal. Table 1 presents the descriptive statistics of our corpus.

**Table 1**. Corpus descriptive statistics.

| **Measure** | **M** | **SD** |
|---|---|---|
| Number of Words (Tokens) | 764.1 | 194.9 |
| Number of Unique Words (Types) | 254.5 | 53.5 |
| Type/Token Ratio | 42.97% | 4.62% |

The SBS analysis provided a ranking of the semantic importance of the key concepts on the three dimensions of connectivity, diversity and prevalence. Results are reported in Figure 1.

**Figure 1. Semantic importance**

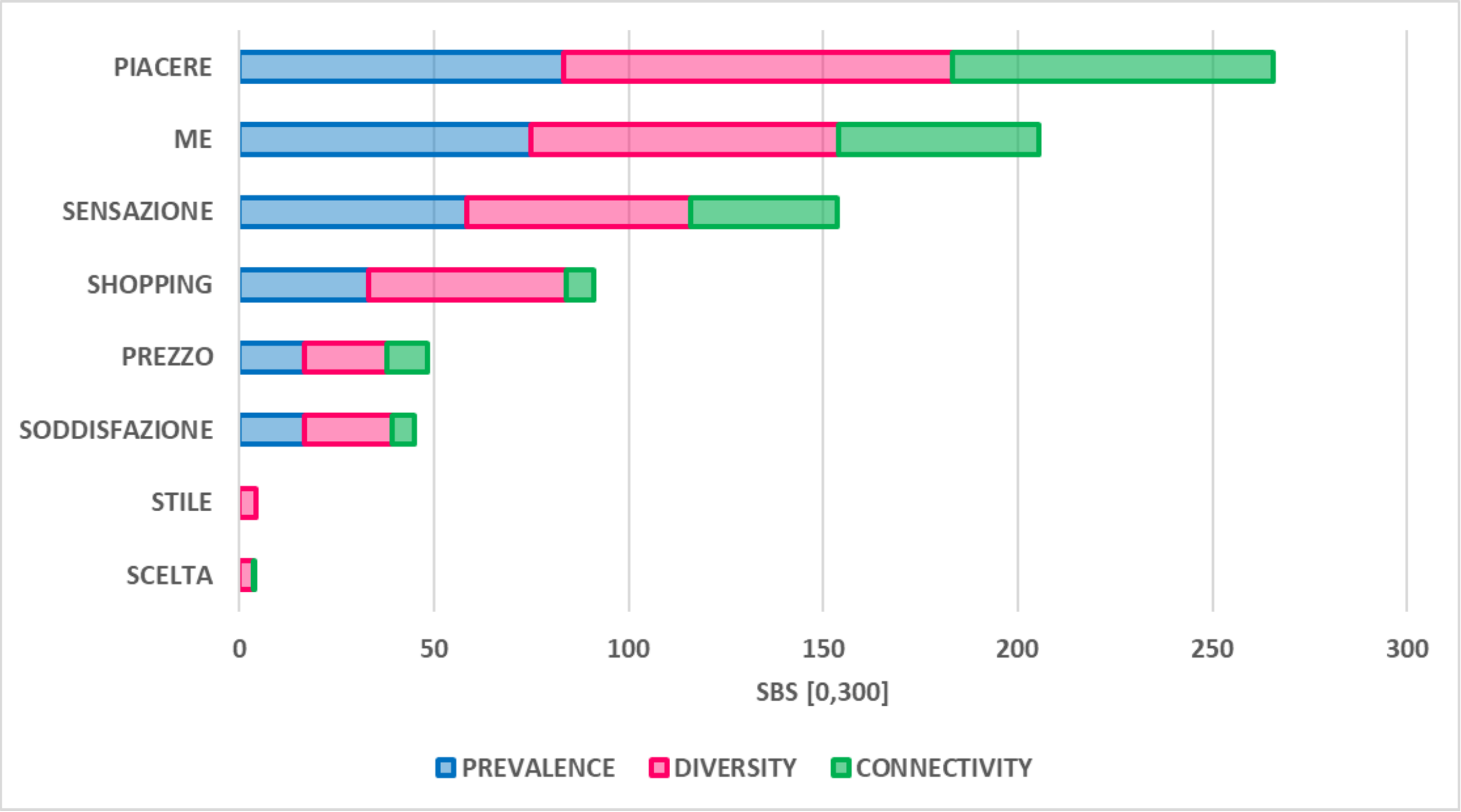

The most relevant concept was *pleasure* (‘piacere’ in Italian), followed by *me/I* (‘me/io’, respectively,

in Italian) and *sensation* (‘sensazione’ in Italian). These three elements resulted as the most frequent (from the calculation of prevalence) and heterogeneous in use (as they have the highest ranking on diversity) and able to connect networks of words (as the connectivity index indicates). This suggests a verbalization of the shopping experience focused on hedonic dimensions related to the self. A second group of words – *shopping*, *price* (‘prezzo’ in Italian) and *satisfaction* (‘soddisfazione’ in Italian) - illuminates a more utilitarian dimension of the verbalized experience that includes an evaluation (e.g., *satisfaction* cf. Oliver, 1980). Finally, and marginally, a third group of words, *style* and *choice* (respectively ‘stile’ and ‘scelta’ in Italian), suggests the presence of a behavioral dimension. As attitude is a three-dimensional mental construct that includes affective, cognitive and behavioral elements, we can conclude that participants verbalized their attitude toward the shopping experience by emphasizing the affective dimension with a clear inward-looking perspective.

This interpretation is further supported by the analysis of the semantic image of each concept. Figure 2 visualizes the most important associations of each concept node.

**Figure 2. Semantic image**

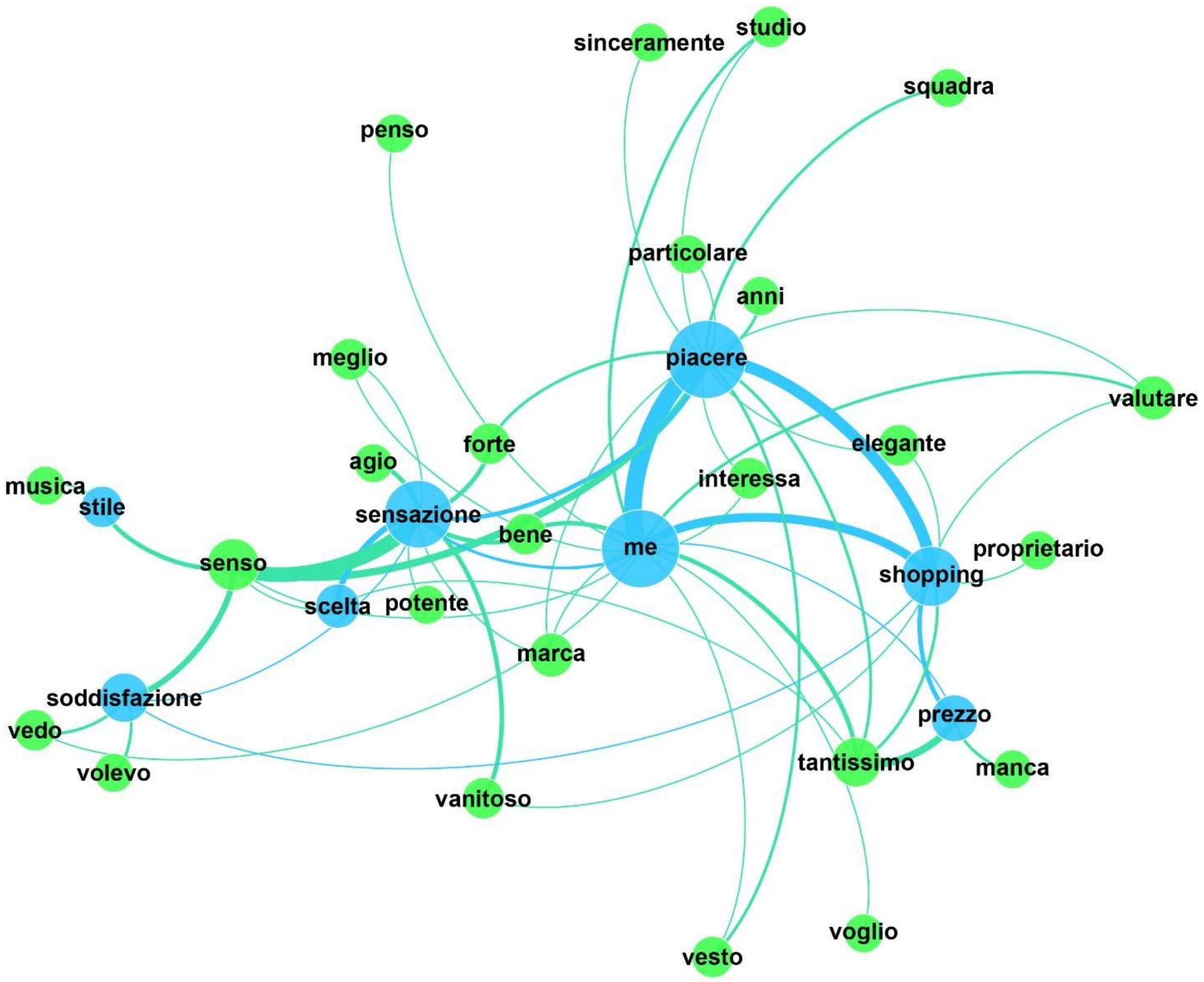

From the words graph, *pleasure* ('piacere' in Italian), *me* ('me' in Italian), and *sensation* ('sensazione' in Italian), *shopping* and *meaning* ('senso' in Italian) represent separate and connected concepts resulting from the shopping experience. Noteworthy, *pleasure*, *shopping* and *me* are strongly connected as well as *sensation* and *meaning*. Overall, this pattern is consistent with a highly arousing experience evaluated from the individual perspective of the self, which provides self-satisfaction as the hedonic gratification resulting from the shopping experience (Matzler et al., 2005). A different concept of *satisfaction* ('soddisfazione' in Italian) results from *I see* ('vedo' in Italian) and *I expected* ('volevo' in Italian), consistent with the Oliver's (1980) conceptualization of transactional satisfaction. *Satisfaction* is weakly connected to *shopping* and *sensation* but is disconnected from *me*. Finally, the two distinct concepts of satisfaction and self-satisfaction are connected through the overall *meaning* (‘senso’ in Italian) of the shopping experience. Indeed, the network of concepts resulting from our data suggests the existence of two distinct but connected forms of satisfaction experienced by customers during shopping. To corroborate this analysis, we studied the similarity of our key concepts in the mind of the interviewees, by looking at their distance in Figure 3.

**Figure 3. Image similarity**

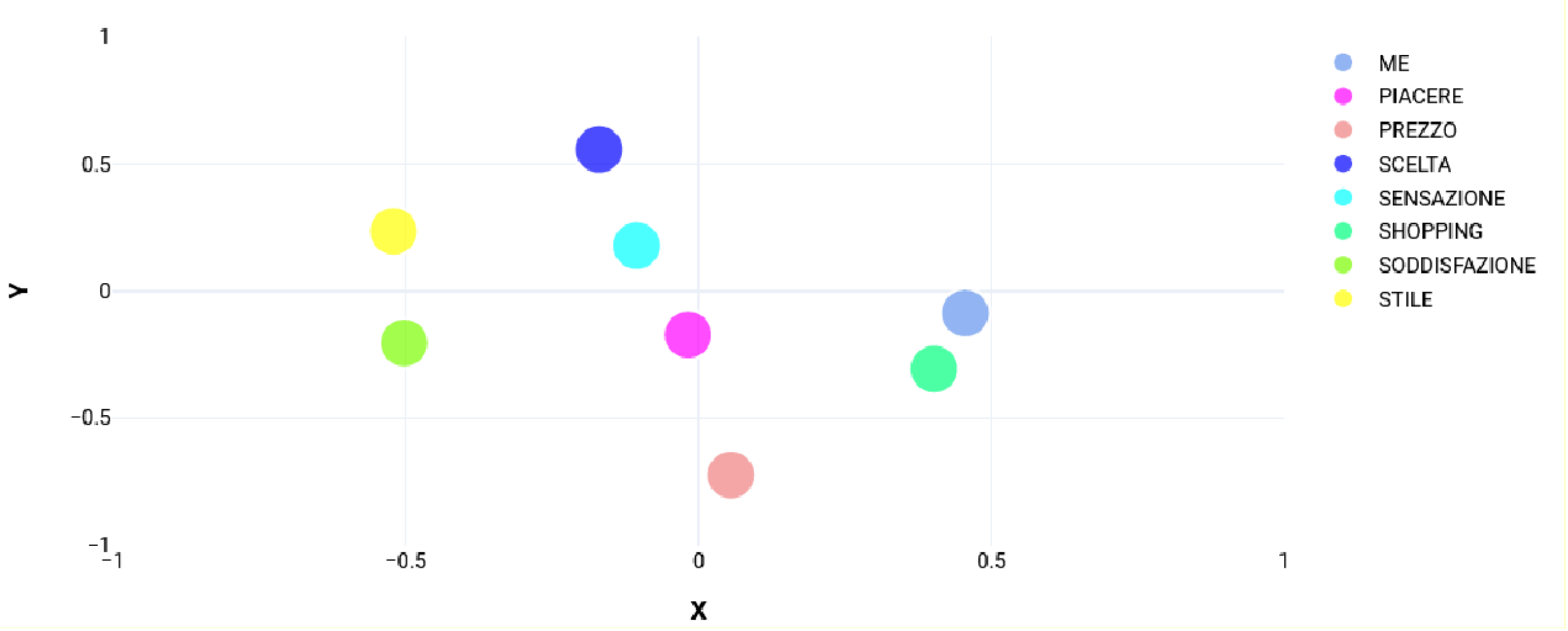

In details, *style* (‘stile’ in Italian) and *satisfaction* (‘soddisfazione’ in Italian) as well *shopping* and *I* or *me* (both ‘me’ in Italian), are the two closest couples of words. At the same time, they are clearly differentiated from the other words, highlighting that the mental representation of individuals translates into a discourse that deeply and spontaneously looks inward to provide a genuine story of their shopping trip. In fact, *price* (‘prezzo’ in Italian) is distant from the other words, in particular from *choice* (‘scelta’ in Italian), which is closer to *sensation* (‘sensazione’ in Italian) and *pleasure* (‘piacere’ in Italian). As a result, it appears that shopping is experienced from the personal perspective of the self with a prominent role of sensations and pleasure. The genuine verbalization of this experience overlooks price and indicates an instinctive dimension.

Lastly, we considered the concepts with the highest connective power (i.e., most central in the discourse), which are *pleasure* (‘piacere’ in Italian), the first person *I* or *me* and the concepts related to arousal, grouped into *sensation* (‘sensazione’ in Italian). Overall, it appears that individuals tend to verbalize their experience through a mental representation of the arousal felt during shopping by choosing those expressions that better constitute a link to their self-identity.

## Discussion

Results show a prominent role of emotions in the discourse about shopping experience and an inward-looking connection to the self. In fact, the focus on the self emerges from the use of the first-person singular pronouns, as suggested by the literature (e.g., Lyons et al., 2018).

The concept of self-satisfaction has been highlighted in terms of emotions and sensations aroused by the shopping trip. The act of shopping arouses not only a sense of (self-)accomplishment or reward, but also a strong connection with and reinforcement of the self-identity. This generates positive emotions that could give a different meaning to shopping activities, more related to self-expressiveness and well-being.

Generally speaking, the results suggest that modern retail environment should enhance the hedonic shopping experience in terms of fun, moods, and emotions. Managerially, this approach will motivate and attract consumers to regularly patronize stores. Retailers could then develop shopping environments that stimulate self-satisfaction and, more widely, the well-being implications of shopping activities and of consumers. Our data show that shopping functional congruence is not explicitly relevant in the textual associations and frequencies but emerges because of the congruence between the aroused emotions and the self.

Although the study suffers some limitations, including the small sample and the population of tattooed people, it opens interesting avenues of research that deserve attention. First, since our study offers an exploratory contribution serving in the context of discovery and not of justification, we focus only on Italian tattooed consumers. Future research should analyze the effects of other elements that potentially could strengthen or weaken the subjective confidence with the self-identity narratives, including cultural elements (e.g., religion, morality). Furthermore, the possibility to compare different individual characteristics, above all cultural identity, on the ability to express shopping experience deserves attention.

Despite the limited sample size, our analyses suggest that listening to customer *outside* the digital world could have relevant and potential implications. In particular, by selecting a population of tattooed people, we strove to exacerbate the self-consciousness of participant and their tendency to visually communicate their personal self-identity to others through their body. Once individuals have the possibility to *talk* about themselves, a tight relationship between their self-identity and the verbalization of their arousal during shopping emerges, in line with previous research (Akpinar and Berger, 2017; Berger et al., 2020; Xu and Zhang, 2018). This conclusion may apply to all people (whether tattooed or not) with different nuances and strength, and deserves further investigation by marketing scholars.

While online textual data offer unprecedented opportunities to analyze the language individuals use to express themselves in consuming contexts (Kübler et al., 2020; Ziemer and Kormaz, 2017), they limit the richness of the language in natural, offline, conditions. In particular, since social media and digital providers are unwilling to allow free access to data after the Cambridge Analytica *affaire*, the literature is teemed with studies on Twitter. Unfortunately, tweets are limited in length and likely to clip individuals' full and genuine self-expression. It is likely that some few words are not enough to express self-identity, experience verbalization and their links. Our study, instead, suggests that research should return to let people talk about themselves without time or space constraints, providing really rich information on their personality, shopping experience and the ways to express them. In this perspective, a fruitful avenue of research could be the analysis of the transcripts of the calls to customer service.